\documentclass[10pt, a4paper]{article}
\usepackage[]{lrec-coling2024} 
\usepackage{cite}
\usepackage{multirow}
\usepackage{amsmath}

\begin{document}

\title{Deep Learning Based Named Entity Recognition Models for Recipes}
%\title{Deep-learning based Named Entity Recognition in Recipes}
%\title{Named Entity Recognition in Recipes}
%\title{Models for Named Entity Recognition in Recipes}
%\title{Deep-learning Models for Named Entity Recognition in Recipes}
%\title{Deep-learning Models for Named Entity Recognition in Recipe Ingredient Phrases}
%\title{Deep-learning Models for Named Entity Recognition in Recipe Text}
%\title{Deep-learning Models for Named Entity Recognition of Ingredient Phrases in Recipes}
%\title{Deep-learning Models for Named Entity Recognition in Recipes}
%\title{Deep-learning based Ingredient Named Entity Recognition (DINNER)}

\name{Mansi Goel$^{1,2,\dagger}$, Ayush Agarwal$^{3,\dagger}$, Shubham Agrawal$^{3,\dagger}$, Janak Kapuriya$^{3,\dagger}$, \\{\bf \large Akhil Vamshi Konam$^{3,\dagger}$, Rishabh Gupta$^{3}$, Shrey Rastogi$^{3}$,} \\{\bf \large Niharika$^{3}$ and Ganesh Bagler$^{1,2}$ }}

\address{$^{1}$Infosys Centre of Artificial Intelligence, IIIT-Delhi\\ $^{2}$Department of Computational Biology, IIIT-Delhi\\ $^{3}$Department of Computer Science, IIIT-Delhi\\ \{mansig, ayush22095, shubham22124, kapuriya22032, \\ konam20513, rishabh21070, shrey21145, niharika21132, bagler\}@iiitd.ac.in}

\abstract{
% 150 to 200 words 
Food touches our lives through various endeavors, including flavor, nourishment, health, and sustainability. Recipes are cultural capsules transmitted across generations via unstructured text. Automated protocols for recognizing named entities, the building blocks of recipe text, are of immense value for various applications ranging from information extraction to novel recipe generation. Named entity recognition is a technique for extracting information from unstructured or semi-structured data with known labels. Starting with manually-annotated data of 6,611 ingredient phrases, we created an augmented dataset of 26,445 phrases cumulatively. Simultaneously, we systematically cleaned and analyzed ingredient phrases from RecipeDB, the gold-standard recipe data repository, and annotated them using the Stanford NER. Based on the analysis, we sampled a subset of 88,526 phrases using a clustering-based approach while preserving the diversity to create the machine-annotated dataset. A thorough investigation of NER approaches on these three datasets involving statistical, fine-tuning of deep learning-based language models and few-shot prompting on large language models (LLMs) provides deep insights. We conclude that few-shot prompting on LLMs has abysmal performance, whereas the fine-tuned spaCy-transformer emerges as the best model with macro-F1 scores of 95.9\%, 96.04\%, and 95.71\% for the manually-annotated, augmented, and machine-annotated datasets, respectively. 
\\ \newline  
}

\maketitleabstract

\def\thefootnote{$\dagger$}\footnotetext{These authors contributed equally to this work.}

%%%%%%%%%%%%%%%%%%%%%%%%%%%%%%%%%%%%%%%%%
\section{Introduction}
Food plays a central role in our lives. Beyond its primary purpose of nourishment and taste, it encompasses a broad spectrum of endeavors touching on health and sustainability. In the modern culinary landscape, where food is not just sustenance but reflects our diverse tastes and interests, information extraction in food texts has become increasingly crucial. As we explore culinary experiences and adapt to dietary preferences, extracting valuable information from food-related texts empowers us to make informed choices. Information extraction~\citep{wei2023zero} enables efficient utilization of food-related data. This includes identifying ingredients and nutritional details in recipes~\citep{kalra2020nutritional}, ensuring dietary safety by detecting allergens~\citep{pellegrini2021exploiting}, optimizing restaurant operations through menu analysis~\citep{syed2021menuner}, enhancing food safety by tracking recalls, cost and sustainability. These technological enhancements provide deeper perspectives on what we eat and facilitate personalized meal planning, culinary research, and innovation in the food industry. 

\noindent Recipes are unstructured text, and named entities are their building blocks. Named entity recognition (NER) is a technique for extracting information from unstructured or semi-structured data with known labels~\citep{krishnan2005named}. It requires the many-to-one mapping of various named entities in text to their domain-specific categories. NER can extract information from various domains, including reviews, news articles, scientific literature, and food texts. NER not only acts as a standalone tool for information extraction but also plays an essential role in a variety of natural language processing (NLP) applications such as text understanding~\citep{zhang2019ernie,cheng2020attending}, information retrieval~\citep{guo2009named,petkova2007proximity}, automatic text summarization~\citep{aone1999trainable}, question answering~\citep{molla2006named}, machine translation~\citep{babych2003improving}, and knowledge base construction~\citep{etzioni2005unsupervised}, etc. Recent studies have implemented various deep-learning models, such as BERT~\citep{liu2023,fang2023,suleman2023floods}, DistilBERT~\citep{Sanh2019,hossain2022,silalahi2022}, DistilRoBERTa~\citep{davidson2021improved,qu2022distantly,rodrigues2022natural}, spaCy~\citep{kumar2023algorithm}, and  flair~\citep{mathis2022extracting,pathak2022asner,kumar4375613manubert}. 

\begin{figure*}[!ht]
\begin{center}
\includegraphics[width=\textwidth]{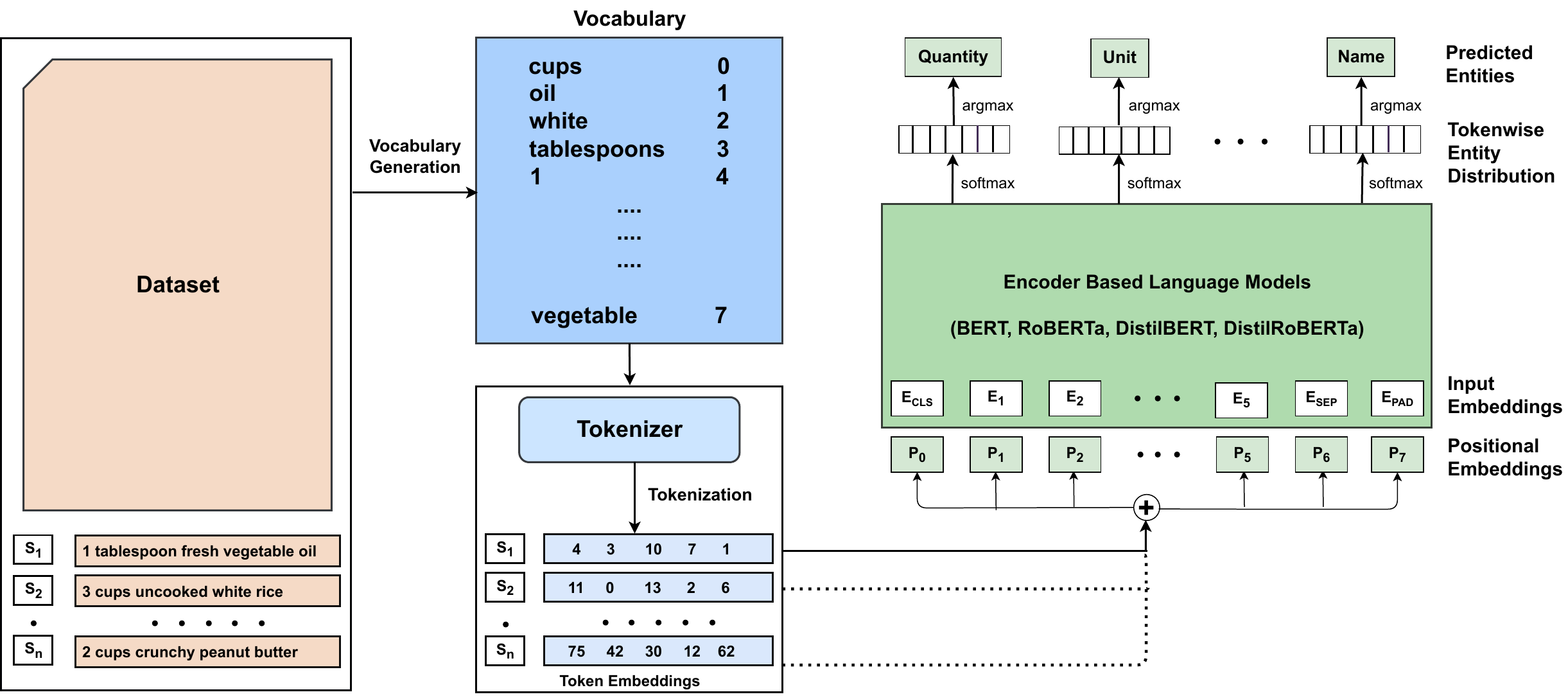} 
\caption{The pipeline implemented for fine-tuning supervised deep learning-based named entity recognition comprises three phases. To begin with, we built vocabularies for each of the three datasets. Further, we utilized these vocabularies to convert every word of an input sentence into corresponding token indexes that were subsequently converted to token embeddings via tokenization. Finally, Encoder-Only language models were employed to predict entity tags for the input token embeddings. The spaCy-transformer emerged as the best model with F1 scores of 95.9\%, 96.04\%, and 95.71\% for Manually\_Annotated, Augmented, and Machine\_Annotated datasets, respectively.}
\label{fig:NER_pipieline}
\end{center}
\end{figure*}

\noindent Traditional NER models, such as Hidden Markov models~\citep{rabiner1986introduction} and conditional random fields~\citep{lafferty2001conditional}, rely heavily on rule-based features~\citep{luo2015joint,passos2014lexicon}. DrNER~\citep{eftimov2017rule} in rule-based NER that can extract food entities from evidence-based dietary recommendations. This work was extended to develop another rule-based NER FoodIE~\citep{popovski2019foodie}, where the rules incorporate computational linguistics information. FoodIE achieved promising results on independent benchmark datasets and has been used to create the FoodBase corpus, the first NER corpus in the food domain. The limitation of the FoodIE method is its dependency on external resources, which have become inaccessible after its publication, rendering the method unusable. With a similar spirit, a data-driven method to find named entities, BuTTER~\citep{cenikj2020butter}, was trained on the FoodBase corpus based on Bidirectional Long Short-Term Memory and conditional random field methods. ~\citet{Radu2022} implemented NER on cooking instructions from multilingual recipes (French, German, and English). They implemented a Conditional Random Field layer on top of Bidirectional Long-Short Term Memory models, achieving F1 scores over 96\% in mono and multi-lingual contexts for all classes. Another research~\citep{brahma2020} implemented a NER approach to identify the food quality descriptors from chats between customers and customer support staff. Previous research~\citep{Diwan2020a} used the RecipeDB dataset~\citep{Batra2020} to identify the named entities in ingredient phrases and cooking instructions. They reported an F1 score of 0.95 (ingredient), 0.88 (processes), and 0.90 (utensils). SciFoodNER~\citep{cenikj2022scifoodner} is a BERT-based method for recognizing named entities in scientific texts and achieved an F1 score of 0.90. NER can accurately identify ingredient names, quantities, unit, state, size, dry/fresh, and temperature within recipes and food-related content. 

\noindent Computational Gastronomy represents the study of food, flavors, nutrition, health, and sustainability from the computing perspectives~\citep{Goel2022}. This new data science niche dramatically changes the outlook on food and cooking, traditionally considered artistic endeavors. In this context, building NER models for recipe texts is an exciting proposition, given its applications spanning multiple domains, including disease prediction, cost estimation, flavor profiling, and comprehensive nutritional analysis of recipes. Herein, we present a computational pipeline by utilizing encoder-based language models to extract NERs from recipe text (Figure~\ref{fig:NER_pipieline}).

\noindent The salient contributions of research studies presented here are (a) the introduction of augmented and machine-annotated ingredient phrase datasets, (b) analysis of the distribution of RecipeDB ingredient phrases, and  (c) a thorough investigation of NER approaches on recipe texts involving statistical, deep-learning-based fine-tuning of language models and few-shot prompting on LLMs.

%%%%%%%%%%%%%%%%%%%%%%%%%%%%%%%%%%%%%%%%%
%%%\section{Materials and Methods}
\section{Dataset}
We have used the manually annotated data consisting of 6,611 ingredient phrases~\citep{Diwan2020a} that were sourced from RecipeDB~\citep{Batra2020}, where all named entities were manually labeled (Manually\_Annotated\_Dataset). An augmented dataset comprising 26,445 ingredient phrases was created by label-wise token replacement, synonym replacement, and shuffling with segments (Augmented\_Dataset).  

\begin{figure*}[!ht]
\begin{center}
\includegraphics[width=0.9\textwidth]{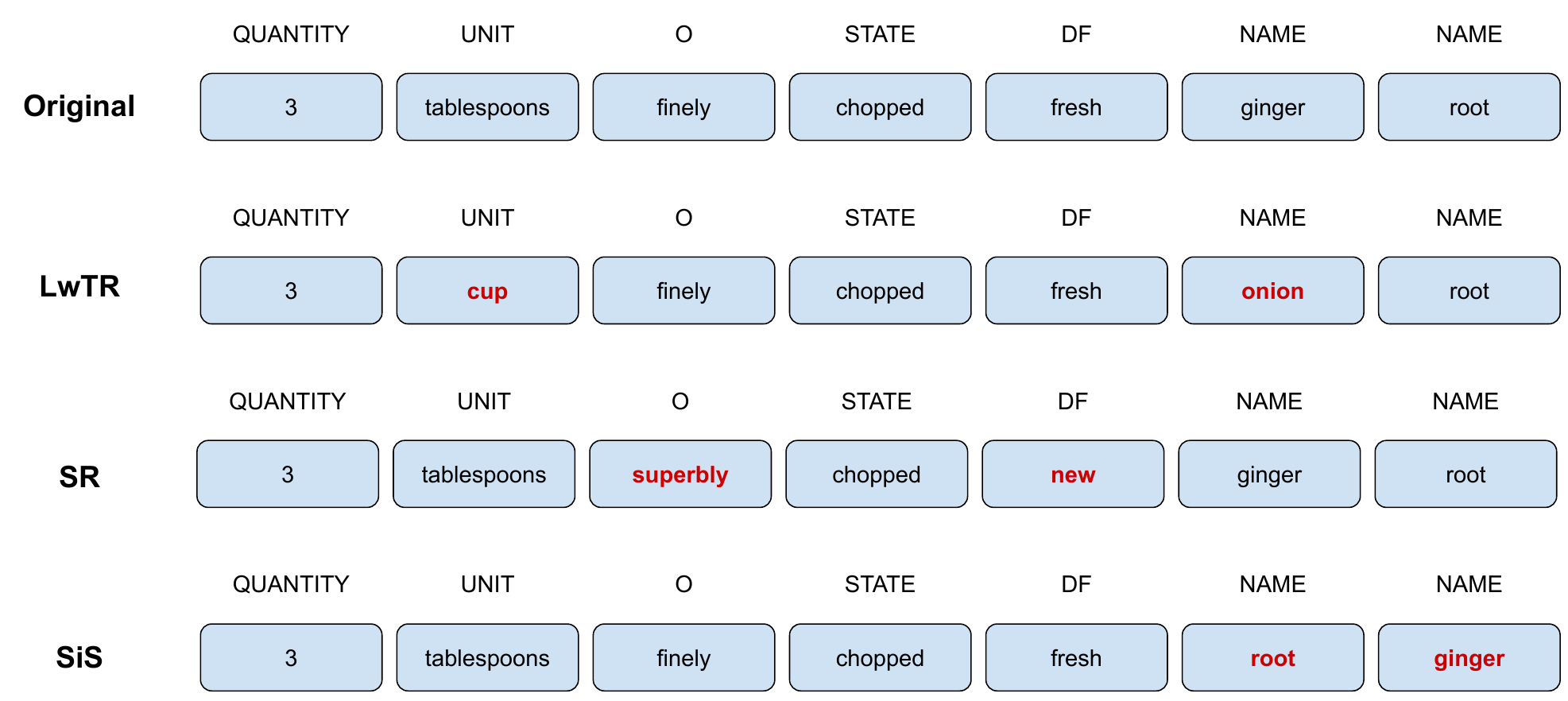} 
\caption{Illustration of Data Augmentation strategies to generate new samples. (a) LwTR: Labelwise Token Replacement: replace a token with a random token of the same label. (b) SR: Synonym Replacement: replace a token with its synonym from Wordnet. (c) SiS: Shuffle within segments: shuffle the tokens under their corresponding label within an ingredient phrase. }
\label{fig:data_aug}
\end{center}
\end{figure*} 

\noindent We created an extensive repository of 349,762 unique ingredient phrases from the RecipeDB dataset~\citep{Batra2020} involving semi-automated processing protocol and human curation (Machine\_Annotated\_Dataset). These ingredient phrases were divided into 2,067 clusters (Stratified Entity Frequency Sampling) based on seven named entity tags (name, quantity, unit, df (dry/fresh), state, size, and temp) and 25\% of data (88,526 phrases) were sampled for training. We used 2,187 Manually\_Annotated\_Dataset for testing. 

\subsection{Data Preprocessing}
Starting with the 1,150,000 ingredient phrases obtained from RecipeDB dataset~\citep{Batra2020}, we implemented a preprocessing protocol of lemmatization and manual annotations. 
A team of culinary experts manually identified the most frequent error patterns present in the dataset (see~\ref{sec:cleaning_appendix}). These mistakes were collectively rectified using Python scripts. 

%%%%%%%%%%%%%%%%%%%%%%%%%%%%%%%%%%%%%%%%%
\subsection{Data Augmentation}
Language models need a larger dataset for training. Hence, to extend the Manually\_Annotated\_Dataset, we implemented three augmentation techniques (Figure~\ref{fig:data_aug}). 

\noindent \textbf{Labelwise Token Replacement (LwTR)}: LwTR replaces the token with a random token from the training set with the same label after taking a call on whether a token should be replaced based on the binomial distribution. This procedure ensures that the original label sequence is preserved.

\noindent \textbf{Synonym Replacement (SR)}: In a procedure analogous to LwTR, the SR method replaces the token randomly with its synonyms from the Wordnet lexical database. 

\noindent \textbf{Shuffle within Segments (SiS)}: In SiS, the token sequence is first split into segments with the same label, so each segment has some probability of shuffling (as per binomial distribution). The token within the same segment is then shuffled while keeping the order of tokens unchanged.

%%%%%%%%%%%%%%%%%%%%%%%%%%%%%%%%%%%%%%%%%
\subsection{Machine-Annotated Dataset}
We had a training dataset with 6,611 and 2,187 labeled ingredient phrases for training and testing. Given ten ingredients per recipe on average in a recipe, this yields around 661 recipes for training and 218 recipes for testing. These data are of limited utility when training transformer-based language models on which our experiments are based and which are known to excel in NLP tasks such as named entity recognition.  

\subsubsection*{Dataset Creation}
Given the size of the ingredient phrase corpus (1,150,000 ingredient phrases), it was deemed impractical to annotate the entire RecipeDB. After removing duplicates (an ingredient phrase may be a part of several recipes), we were left with 349,762 unique phrases. We adopted a hybrid approach to address this challenge. First, we trained the Stanford NER on the labeled corpus (6,611 + 2,187 = 8,798 Ingredients) to annotate the unique ingredient phrases from RecipeDB. Then, we manually cleaned the machine-generated annotations to identify the error patterns and correct them programmatically. We implemented Stratified Entity Frequency Sampling, a clustering and sampling approach, to sample 25\% (88,526 phrases) of the unique ingredient phrases. 

\subsubsection*{Stratified Entity Frequency Sampling}
The unique challenges posed by our dataset led to the development of a clustering and sampling technique that we term `Stratified Entity Frequency Sampling (SEFS).' SEFS ensures a diverse and representative selection of annotated data from a vast corpus, maximizing the capture of varied ingredient phrase patterns.

\noindent SEFS operates on the premise that ingredient phrases vary based on the combination and frequency of entities they contain. Some phrases may contain the ingredient's name only, while others could be more descriptive, indicating quantity, unit, state, size, and temperature. Ensuring a wide-ranging representation of these combinations in our sample was imperative to train a robust model.

\noindent \textbf{Clustering} The first step in SEFS is to cluster the unique ingredient phrases based on their entity composition. An entity frequency vector is created for each phrase, where each vector component represents the count of a specific entity (name, quantity, unit, state, size, or temperature) in the phrase. These vectors serve as the basis for clustering, where each unique vector corresponds to a cluster. This ensures that ingredient phrases with the same entity composition and frequency are grouped.

\noindent \textbf{Sampling} Once clustered, we sample from these groups to create our dataset. A uniform sampling might not capture the richness and variability of the corpus. Therefore, we adopt a stratified sampling approach. In this method, we sample a fixed proportion (25\%, in our case) from each cluster. This guarantees that the resultant dataset contains diverse ingredient phrase patterns.

\begin{figure*}[!ht]
\begin{center}
\includegraphics[width=0.9\linewidth]{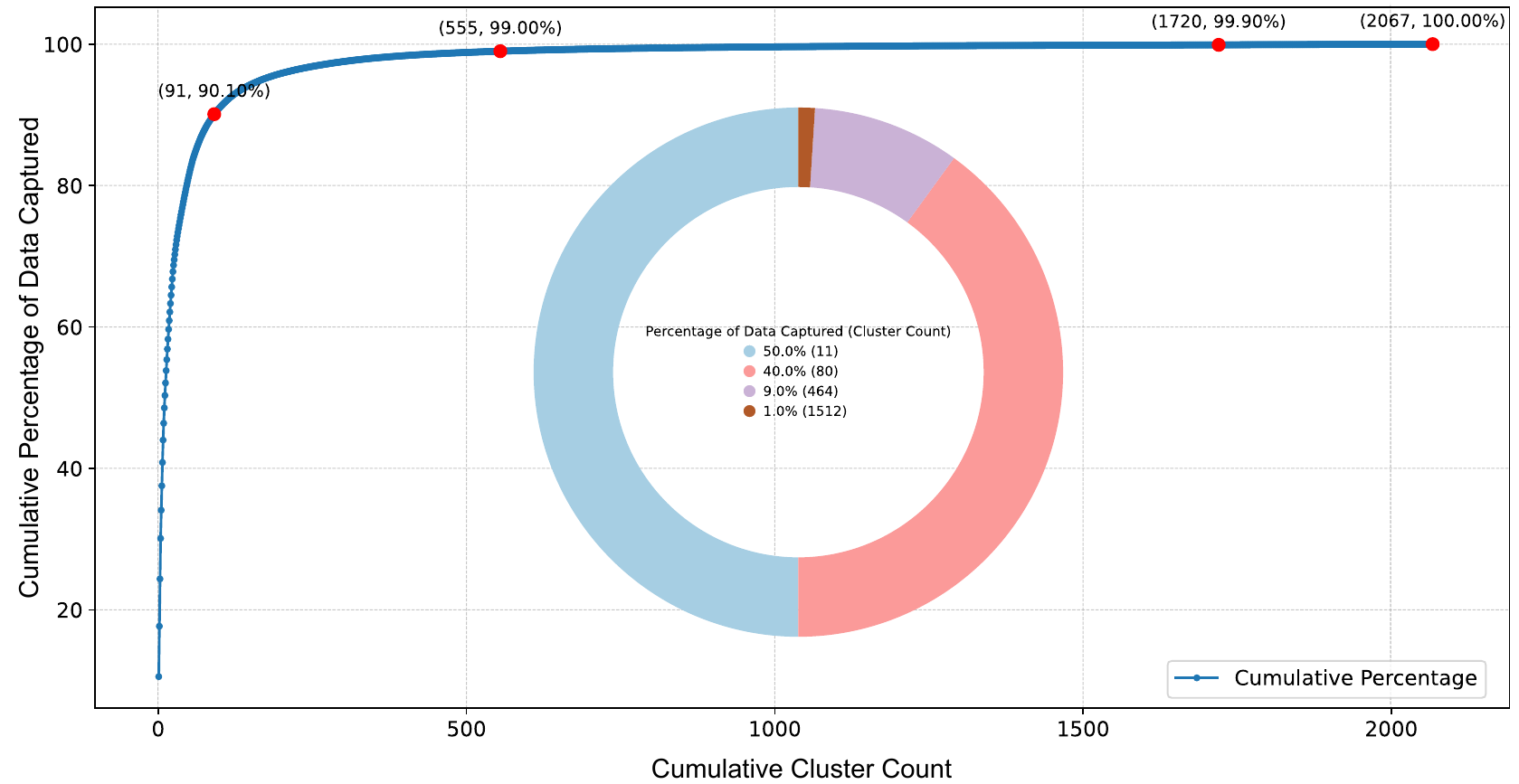} 
\caption{Analysis of the percentage of ingredient phrases captured by various clusters. The distribution is extremely skewed, with a few clusters hoarding most ingredient phrases. Half of the ingredient phrases, for example, are captured by merely the eleven largest clusters. }
\label{fig:data_cluster}
\end{center}
\end{figure*} 

\noindent SEFS ensures that our sample is not biased towards any particular type of ingredient phrase. It captures the breadth and diversity of the RecipeDB, making it particularly suited for training transformer-based models that thrive on varied data. Moreover, the stratified sampling ensures that even rarer patterns, which could be missed in a random sampling approach, are included in the dataset.

\noindent Figure~\ref{fig:data_cluster} depicts the skewed distribution of ingredient phrases across clusters. Around 90\% of the total ingredient phrases (1.15 million) can be represented by only 91 unique entity frequency vectors, and the remaining 10\% of the phrases require 1,976 different frequency vectors for their representation. This shows that random sampling of ingredient phrases may lead to a bias towards the majority frequency vectors and justifies the SEFS sampling strategy.

%%%%%%%%%%%%%%%%%%%%%%%%%%%%%%%%%%%%%%%%%
\section{Named Entity Recognition Models}
\subsection{Model Configurations}
Building upon the previous work~\citep{Diwan2020a}, we re-implemented the Stanford-NER~\citep{finkel-etal-2005-incorporating}. The Standford NER was trained using CRFClassifier with default parameters on an 8 GB CPU RAM system. We implemented diverse deep-learning NER models (BERT, DistilBERT, RoBERTa, and DistilRoBERTa) and NLP frameworks (spaCy, and flair) to find the named entities in the ingredients section. We fine-tuned our datasets on base-case variants of BERT, DistilBERT, RoBERTa, and DistilRoBERTa models with their pre-trained weights using an SGD optimizer with a learning rate 1e-2. All these models were run on an NVIDIA A100 80GB PCIe GPU card with a batch size of 44 and up to 12 epochs. We have used two different pipelines of spaCy 3.6.1 (en\_core\_web\_lg - a classical rule-based NLP pipeline optimized for CPU, and en\_core\_web\_trf - a RoBERTa-based transformer pipeline). Flair used a pre-trained xlm-roberta-large model to perform the NER. 

\subsection{Modelling Techniques}
BERT~\citep{Toutanova2019} captures the contextual nuances in language by considering the surrounding context of a word in a sentence. Apart from BERT, we employed its other three variants - DistilBERT~\citep{Sanh2019}, RoBERTa~\citep{liu2023} and DistilRoBERTa~\citep{Sanh2019}. NLP frameworks such as spaCy~\citep{Honnibal_spaCy_2020}, flair~\citep{akbik-etal-2019-flair} have been implemented to find the named entities of ingredient phrases. A tool Stanford NER~\citep{finkel-etal-2005-incorporating} employs Conditional Random Fields to analyze and tag entities in a given text with their respective categories. One of its notable features is its ability to recognize and classify entities in multiple languages, making it valuable for multilingual applications.

%%%%% TABLE: Stanford NER Implementation and Comparison
\begin{table*}[]
\centering
\begin{tabular}{|c|cccccc|}
\hline
\multirow{3}{*}{\textbf{Test Set}} &
  \multicolumn{6}{c|}{\textbf{Train Set}} \\ \cline{2-7} 
 &
  \multicolumn{3}{c|}{\textbf{\citet{Diwan2020a}}} &
  \multicolumn{3}{c|}{\textbf{Present Study}} \\ \cline{2-7} 
 &
  \multicolumn{1}{c|}{AR} &
  \multicolumn{1}{c|}{GK} &
  \multicolumn{1}{c|}{Both} &
  \multicolumn{1}{c|}{AR} &
  \multicolumn{1}{c|}{GK} &
  Both \\ \hline
AR &
  \multicolumn{1}{c|}{96.82} &
  \multicolumn{1}{c|}{93.17} &
  \multicolumn{1}{c|}{97.09} &
  \multicolumn{1}{c|}{96.82} &
  \multicolumn{1}{c|}{93.31} &
  97.04 \\ \hline
GK &
  \multicolumn{1}{c|}{86.72} &
  \multicolumn{1}{c|}{95.19} &
  \multicolumn{1}{c|}{94.98} &
  \multicolumn{1}{c|}{86.71} &
  \multicolumn{1}{c|}{95.16} &
  95.02 \\ \hline
Both &
  \multicolumn{1}{c|}{89.72} &
  \multicolumn{1}{c|}{94.72} &
  \multicolumn{1}{c|}{96.11} &
  \multicolumn{1}{c|}{\textbf{89.16}} &
  \multicolumn{1}{c|}{94.72} &
  \textbf{95.52} \\ \hline
\end{tabular}
\caption{Performance comparison of \citet{Diwan2020a} and our implementation of Stanford NER. AllRecipes.com (AR) and geniuskitchen.com (GK) refer to the source of recipes from where the raw data was compiled to create the Manually\_Annotated dataset.}
\label{tab:stanford-ner-implement-compare}
\end{table*}

%%%%%% FIGURE: NER MAIN RESULTS - f1 && lOSS
\begin{figure*}[!ht]
\begin{center}
\includegraphics[width=\textwidth]{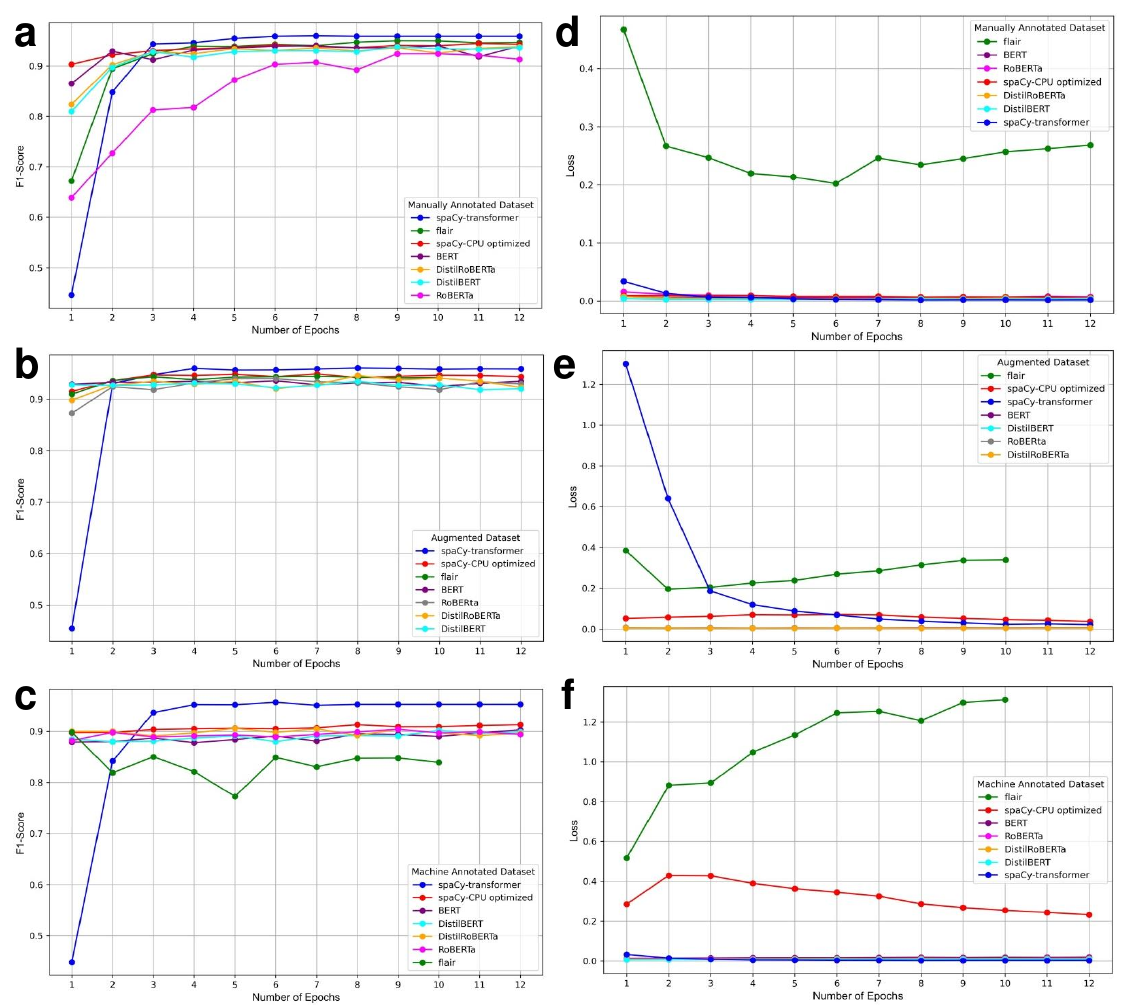} 
\caption{Model Comparison based on F1-scores and Loss. (a), (b) and (c) represent epoch-wise F1-score for Manually Annotated, Augmented, and Machine Annotated Datasets, respectively. Similarly, (d), (e) and (f) represent the epoch-wise Loss score for three datasets.}
\label{fig:ner_fl_loss}
\end{center}
\end{figure*}

%%%%%%%%%%%%%%%%%%%%%%%%%%%%%%%%%%%%%%%%%
\section{Model Evaluation}
We employ macro-F1 score, precision, and recall to evaluate our models' predictive performance. These metrics address the inherent class imbalance in our datasets, where accuracy can be misleading. The F1 score provides a robust measure in such cases.  Precision and recall are equally critical for our task, as we prioritize correctly identifying all valid ingredient tags (particularly names and quantities) without omissions.  While the macro-F1 score is an average of tag-wise F1 scores, it's important to note that it doesn't directly follow the typical harmonic mean relationship with precision and recall. This is because macro-averaging calculates these metrics separately for each label and then averages them, giving equal weight to all labels – a crucial distinction for interpreting results in multi-label classification tasks.

% \begin{equation} \label{avgF1}
%     \text{Avg. F1} = \frac{2 \times \text{Avg. Precision} \times \text{Avg. Recall} }{\text{Avg. Precision} + \text{Avg. Recall}}
% \end{equation}

%%%%%%%%%%%%%%%%%%%%%%%%%%%%%%%%%%%%%%%%%
\section{Results}
Pattern recognition aimed at NER across manual, augmented, and machine-annotated datasets is a difficult task due to degenerate tags corresponding to the same named entity. These ambiguous associations have origins in the linguistic subtleties referring to food's taste, value, and utility. For example, the word `sour’ in `sour cream’ signifies STATE, whereas in `ice cream,' it collectively represents an ingredient; hence, both entities should belong to the NAME tag.

\noindent Herein, we present state-of-the-art models based on deep learning and statistical approaches for named entity recognition in recipe texts. This section is arranged as follows: Section~\ref{sec:stanford-ner} discusses the implementation of Stanford NER~\citep{finkel-etal-2005-incorporating}, which uses statistical-based techniques for NER. In Section~\ref{sec:deep-learning}, we evaluate relevant deep-learning-based models fine-tuned on our datasets for performance. Section~\ref{sec:tag-wise-analysis} describes the tag-wise analysis of named entities using the best performing model, and finally, Section~\ref{sec:few-shots-analysis} delves into the few-shot prompting experiments using state-of-the-art LLMS.

\subsection{\label{sec:stanford-ner} Stanford NER Implementation}
We used the Stanford NER~\citep{finkel-etal-2005-incorporating}, to reproduce the earlier work of Nirav et.al~\citep{Diwan2020a} and have found consistent results (Table~\ref{tab:stanford-ner-implement-compare}). We obtained the same results for seven out of nine experiments, and for the rest of the two, the deviation was <1\%. These results signify the importance of CRF-based methods, which have been the go-to methods for recipe NER in most previous works~\citep{Diwan2020a, patil2020named, wei2016disease, yang2018conditional, sato2017segment}. By building on the learnings from these articles and rooted in extensive datasets introduced in this study, we implement deep-learning-based, state-of-the-art fine-tuned models.

\subsection{\label{sec:deep-learning}Supervised Fine-tuning of Encoder-based Language Models}
To enhance the performance of Named Entity Recognition on recipes, we began with a baseline model, Stanford NER~\citep{finkel-etal-2005-incorporating}. We further implemented seven deep-learning models, including BERT variants (BERT, DistilBERT, RoBERTa, and DistilRoBERTa) and NLP toolkits (SpaCy with CPU optimization, SpaCy equipped with transformer, and flair). To ensure a comprehensive assessment, each model was fine-tuned across three distinct datasets before being consistently evaluated on the Manually Annotated test dataset of 2187 ingredient phrases~\citep{Diwan2020a}. 

%%%%%%% TABLE: PERFORMANCE EVALUATION OF DEEP-LEARNING MODELS ON THREE DATASETS
\begin{table*}[]
\centering
\begin{tabular}{|c|c|c|c|c|c|c|c|c|c|}
\hline
\multicolumn{1}{|c|}{\textbf{Modelling Technique}} &
\multicolumn{3}{|c|}{\textbf{Manually\_Annotated}} &
\multicolumn{3}{|c|}{\textbf{Augmented}} &
\multicolumn{3}{|c|}{\textbf{Machine\_Annotated}} \\
\cline{2-10}
\multicolumn{1}{|c|}{} & \textbf{F1 (\%)} & \textbf{P (\%)} & \textbf{R (\%)} & \textbf{F1 (\%)} & \textbf{P(\%)} & \textbf{R (\%)} & \textbf{F1 (\%)} & \textbf{P(\%)} & \textbf{R (\%)}\\
\hline
\textbf{spaCy-transformer} & \textbf{95.90} & \textbf{95.89} & \textbf{95.91} & \textbf{96.04} & \textbf{96.05} & \textbf{96.04} & \textbf{95.71}  &  \textbf{95.73} & \textbf{95.69} \\
\hline
spaCy-CPU optimized & 94.46 & 94.52 & 94.41 & 94.91 & 94.92 & 94.90 & 91.30 & 91.36 & 91.24 \\
\hline
Stanford NER & 95.52 &  95.64 &  95.39 &  95.16 & 94.37 & 95.96 & 89.9 & 91.31 & 88.53 \\
\hline
DistilBERT  & 93.80 & 95.20 & 93.60  & 93.50 & 93.50 & 94.60 & 90.20 & 92.20 & 89.70\\
\hline
BERT & 94.00 & 94.70 & 94.10 & 93.60 & 93.70 & 94.10 & 90.30 & 91.50 & 90.20 \\
\hline
DistilRoBERTa & 93.80 & 94.80 & 93.90 & 94.60 & 94.10 & 95.90 & 90.60 & 91.60 & 90.60\\
\hline
RoBERTa & 92.40 & 92.90 & 92.60 & 94.00 & 94.50 & 94.10 & 90.40 & 91.60 & 90.20\\
\hline
flair & 95.01 & 96.11 & 96.05 & 94.45 & 95.87 & 96.14 & 89.85 & 88.71 & 89.22\\
\hline
\end{tabular}
\caption{Performance Evaluation on Manually Annotated, Augmented and Machine Annotated Datasets}
\label{tab:performance}
\end{table*}

\noindent Figure~\ref{fig:ner_fl_loss} depicts epoch-wise F1 and validation loss scores for all three datasets across all models. Table~\ref{tab:performance} encapsulates the results from the best epoch for every dataset-model pair. Despite starting with a lower F1 score, the spaCy-transformer exhibits a rapid learning curve, eventually surpassing the performances of its counterparts. Such discrepancies, especially during the initial epochs across various models, can be attributed to the inherent variability arising from the seed values of model weights and consistent hyperparameters employed. The Augmented dataset, as expected, shows slight performance gains, which is explained by the fact that DL models are data-hungry and given more examples, they would learn the entity-tag mapping better. However, the Machine\_Augmented dataset with silver labels created using DL models previously trained on Manually\_Annotated datasets appears to echo the inherent variability and noise, coupled with potential mislabeling. This explains a slight decrease in its performance compared to the manually annotated dataset.

\noindent A particularly captivating observation emerged from our analysis of the Distil-versions compared to their original BERT-based counterparts. Contrary to conventional assumptions, the Distil-variants held their ground and frequently outperformed the base models. This phenomenon merits a closer examination. Several plausible factors could be driving this unexpected outcome. Firstly, the base BERT variants might be predisposed to overfitting the peculiarities of the training set. Such a tendency would culminate in an escalated validation loss, suggestive of an overly tailored model struggling to generalize to new, unseen data.

\noindent Additionally, the presence of fine-grained, spurious correlations within the dataset could be more readily captured by these base models. While seemingly advantageous, this heightened sensitivity might be counterproductive by leading the model to internalize these inconsequential patterns as meaningful, skewing its predictions. Moreover, the potential presence of label noise within the datasets might cause Base BERT models to be overly adept at learning these noise-influenced labels. Consequently, while they might produce tags mirroring the original distribution, these tags might deviate from the expected results in the validation set, thereby being marked erroneous. Summarising, The Distil versions, being smaller with fewer parameters, are weaker in capturing the `bad patterns'--spurious correlations and label noise, which surprisingly acts in favor of their performance metrics.

\noindent As we see from our results on the augmented dataset, some models perform better on the augmented datasets, such as spaCyNER and DistillRoBERTa. Because deep-learning-based language models are data-hungry, we enhanced the volume of our dataset by using data augmentation techniques. Consequently, the model performances get a boost as they get more examples to learn about the inherent nature of ingredient phrases.

\noindent Analysis of the results obtained in the previous section reveals that spaCy-transformer stands out as the best deep-learning-powered package for recognizing entity tags in recipe texts. It outperformed all other models and baselines on all three datasets. It has also shown stable, consistent learning for our models with the least variance compared to others, as shown in Figure~\ref{fig:ner_fl_loss}.

\subsection{\label{sec:tag-wise-analysis} Tag-wise Analysis of Named Entities}
\begin{figure}[!ht]
\begin{center}
\includegraphics[width=\linewidth]{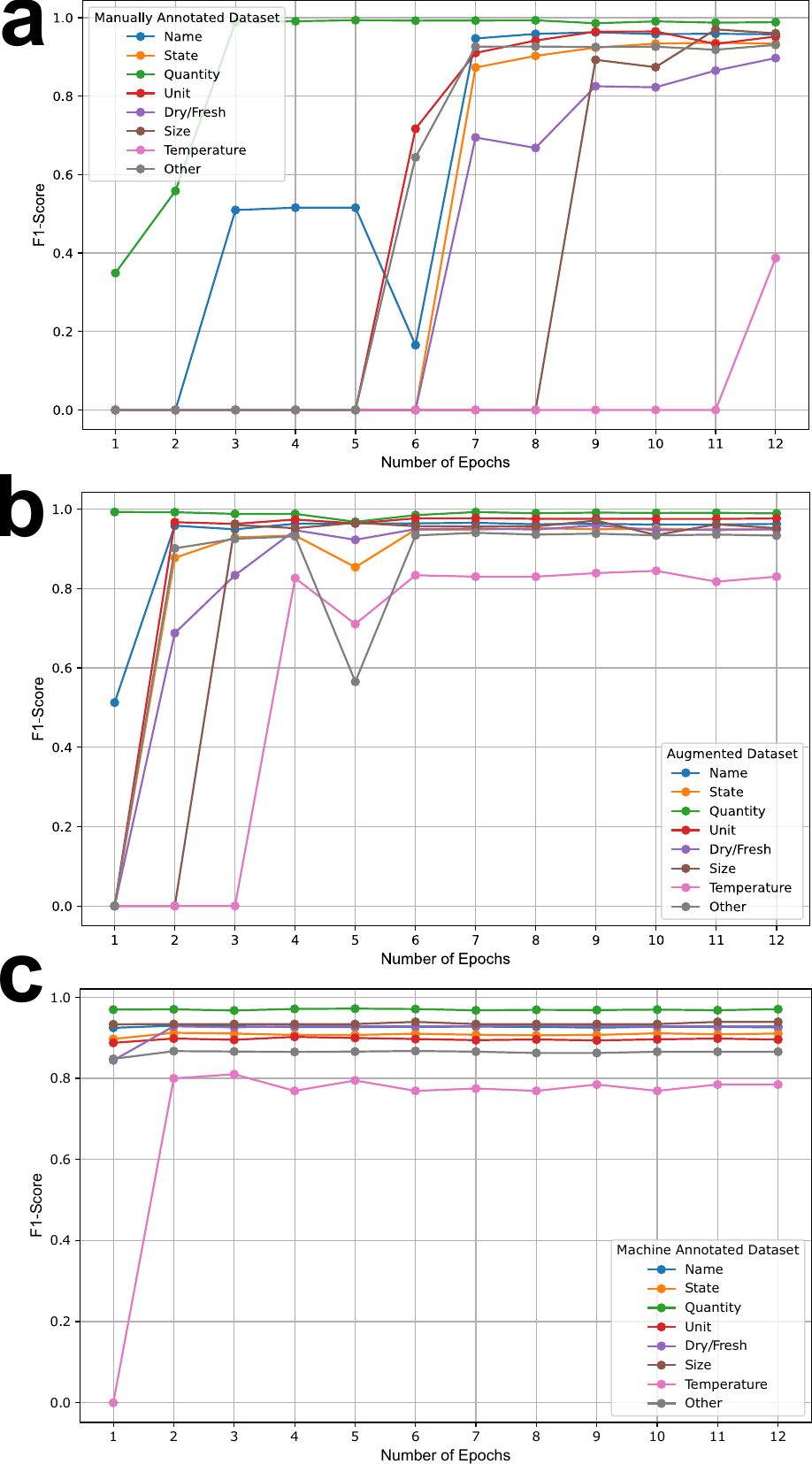} 
\caption{Tag-wise learnability of named entities and their final results using the best-performing model--the spaCy-transformer. Figures (a), (b) and (c) depict these results for the Manually\_Annotated, Augmented, and Machine\_Augmented datasets, respectively.}
\label{fig:classwise-F1-best-model-all-datasets}
\end{center}
\end{figure}

In our investigation of epoch-wise learning trends for various entity tags using our top-performing model, a notable correlation emerges between the frequency of a tag in the dataset and its learning trajectory within the model. Consistently, across all three datasets, the `Quantity' tag exhibits the earliest and most robust learning, while the `Temperature' tag lags, both in initiation and overall learning, as shown in Figure~\ref{fig:classwise-F1-best-model-all-datasets} by their F1 scores. This disparity underscores the model's limitations in grasping rare tags as effectively as with prevalent ones. A plausible interpretation of this observation is that while attempting semantic understanding, the models also rely on memorizing specific entity-tag pairings. Consequently, less frequent tags that offer fewer memorization opportunities tend to be under-learned.

%%%% FIGURE: TAG-WISE ANALYSIS OF NAMED ENTITIES: F1 SCORE 

\subsection{\label{sec:few-shots-analysis} Analysis of Few-Shot Prompting on LLMs}
\begin{figure*}[!ht]
\begin{center}
\includegraphics[width=\linewidth]{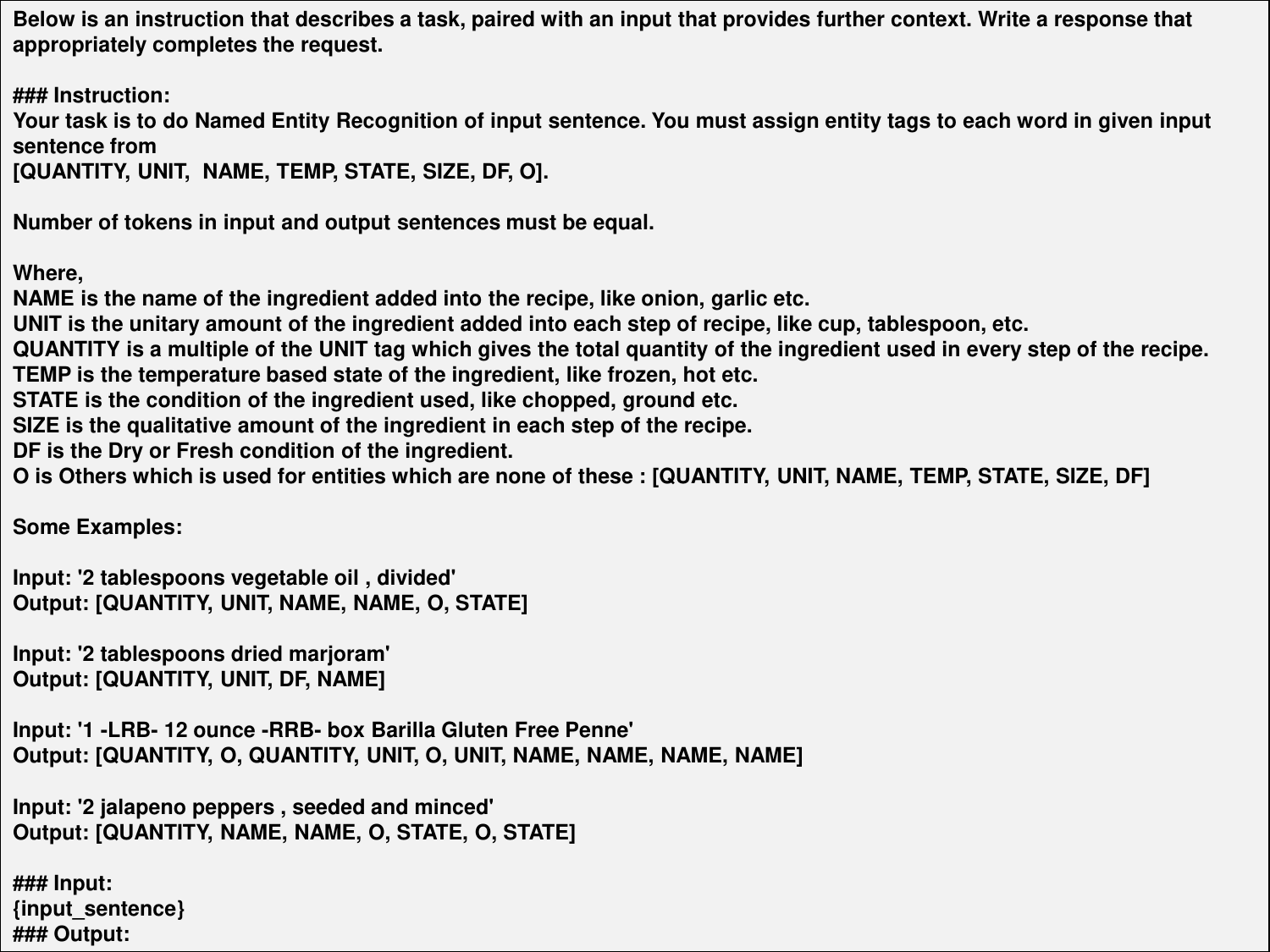} 
\caption{The hand-crafted prompt given to LLMs during Few-Shot Prompting.}
\label{fig:prompt-NER}
\end{center}
\end{figure*}

Few-shot NER leverages the power of LLMs, such as Chat-GPT and GPT-4~\citep{wang2023gpt, ji2023vicunaner}, to tackle the challenging task of entity recognition with minimal annotated data. A prompt is given as input to the LLM that outlines the NER task and specifies the context and available examples (~\ref{fig:prompt-NER}). This prompt acts as a few-shot learning signal, enabling the model to understand the task and context. The pre-trained LLM predicts named entities in a given text. Few-shot NER is useful with limited labeled data, as it can quickly adapt to new entity types and domains. While fine-tuning specific data can further enhance performance, the strength of LLMs lies in their ability to perform remarkably well in a wide range of NLP tasks.

\noindent Table~\ref{tab:prompt} indicates that pre-trained LLMs have limited exposure to food and culinary datasets during their initial pretraining. Consequently, their performance in in-context learning, especially in food-related named entity recognition, is suboptimal. This deficiency in domain-specific knowledge acquired during pretraining significantly affects their in-context learning capabilities and overall task performance. It underscores the need to fine-tune these models with domain-specific datasets to enhance their effectiveness in specialized tasks.

\begin{table}[!ht]
\centering
\begin{tabular}{|c|c|c|}
\hline
\textbf{Model} & \textbf{Macro-F1 (\%)} & \textbf{Micro-F1 (\%)} \\
\hline
LLaMA2-7b  & 5.88  & 44.29 \\
\hline
LLaMA2-13b & 17.06 & 54.20 \\
\hline
Mistral-7b & 32.78 & 47.51 \\
\hline
Vicuna-7b & 32.90 & 51.41 \\
\hline
\end{tabular}
\caption{Results of NER using Few-Shot Prompting on the state-of-the-art LLMs. }
\label{tab:prompt}
\end{table}

%%%%%%%%%%%%%%%%%%%%%%%%%%%%%%%%%%%%%%%%%
\section{Discussion and Conclusions}
Our study presented one of the most extensive labeled data resources of named entities from recipe ingredient phrases. Further, we present deep-learning and statistical models built to achieve state-of-the-art results. Nonetheless, our studies are limited in certain aspects of culinary context, nuances of data, and modeling paradigm.       

\noindent Our present study focuses on only ingredient phrases while not accounting for the recipe instructions, which often carry semantic information about cooking that encodes cultural nuances. Further, static pre-trained models, such as BERT, RoBERTa, and XLM-RoBERTa, come with inherent biases and might not be fine-tuned to capture the nuances of the food lexicon. Complex culinary instructions may not be amenable to extracting meaningful information. For example, the phrase `ground roasted peanuts' holds multiple layers of information, posing a severe challenge for NER. Names of ingredients unique to certain cuisines might be tokenized sub-optimally, leading to NER errors.

\noindent In the future, this research may be extended to include LLM fine-tuning, implementing NERs on cooking instruction, prompt engineering for LLMs for NER on recipes, soft prompt tuning, chain of thought, and implementation of multilingual NER.    

%%%%%%%%%%%%%%%%%%%%%%%%%%%%%%%%%%%%%%%%%
\section{Acknowledgements}
GB thanks Infosys Center of Artificial Intelligence,  Centre of Excellence in Healthcare, and IIIT-Delhi for the computational support. MG thanks IIIT-Delhi for the research fellowship. The authors thank Technology Innovation Hub (TiH) Anubhuti for the research grant. 

\nocite{*}
\section{Bibliographical References}
\bibliographystyle{lrec-coling2024-natbib}
\bibliography{NER}

\appendix

\section{NER Tagging Inference Results}
\label{sec:inference_appendix}

\subsection{Error Analysis Comparing spaCy-transformer with Stanford NER}

\begin{figure*}[]
\begin{center}
\includegraphics[width=0.9\linewidth]{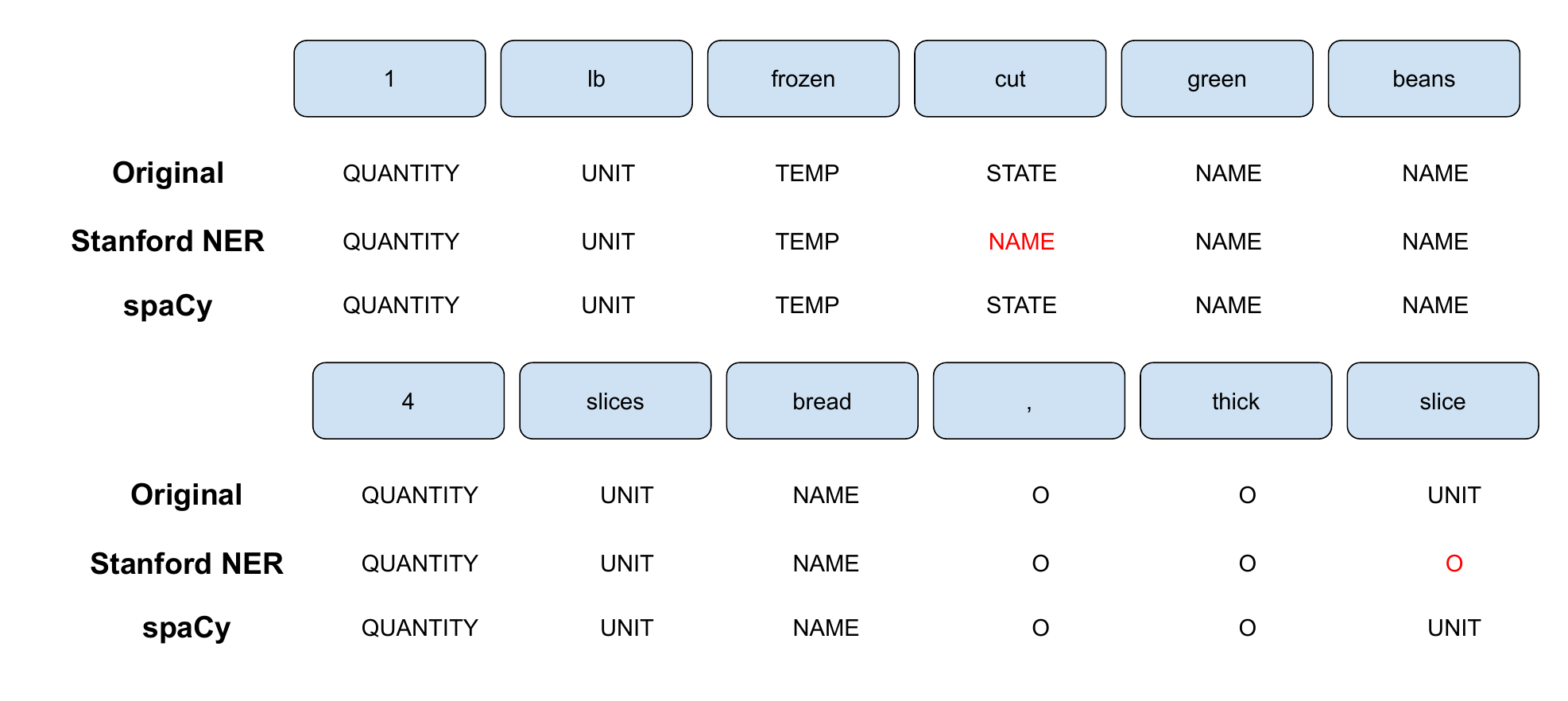} 
\caption{Error Analysis of Stanford NER tagger. Stanford NER tagger incorrectly classifies ``cut” as NAME instead of STATE, which was correctly identified by spaCy-transformer. Similarly, ``slice” classifies as OTHER instead of UNIT, which was correctly identified by spaCy-transformer.}
\label{fig:standard_error}
\end{center}
\end{figure*}

The two most frequent error patterns of Stanford NER that emerged in our analysis were the misclassification of the entity STATE as NAME and the entity UNIT as OTHER (see Figure~\ref{fig:standard_error}). We exemplify these error patterns, showcasing instances where Stanford NER and spaCy-transformer differ in their predictions.

This study reveals a significant breakthrough: spaCy-transformer outperforms the established Stanford NER tagger in recipe entity classification tasks.

\subsection{Erroneous Predictions using spaCy-transformer}
\begin{figure*}[]
\begin{center}
\includegraphics[width=0.9\linewidth]{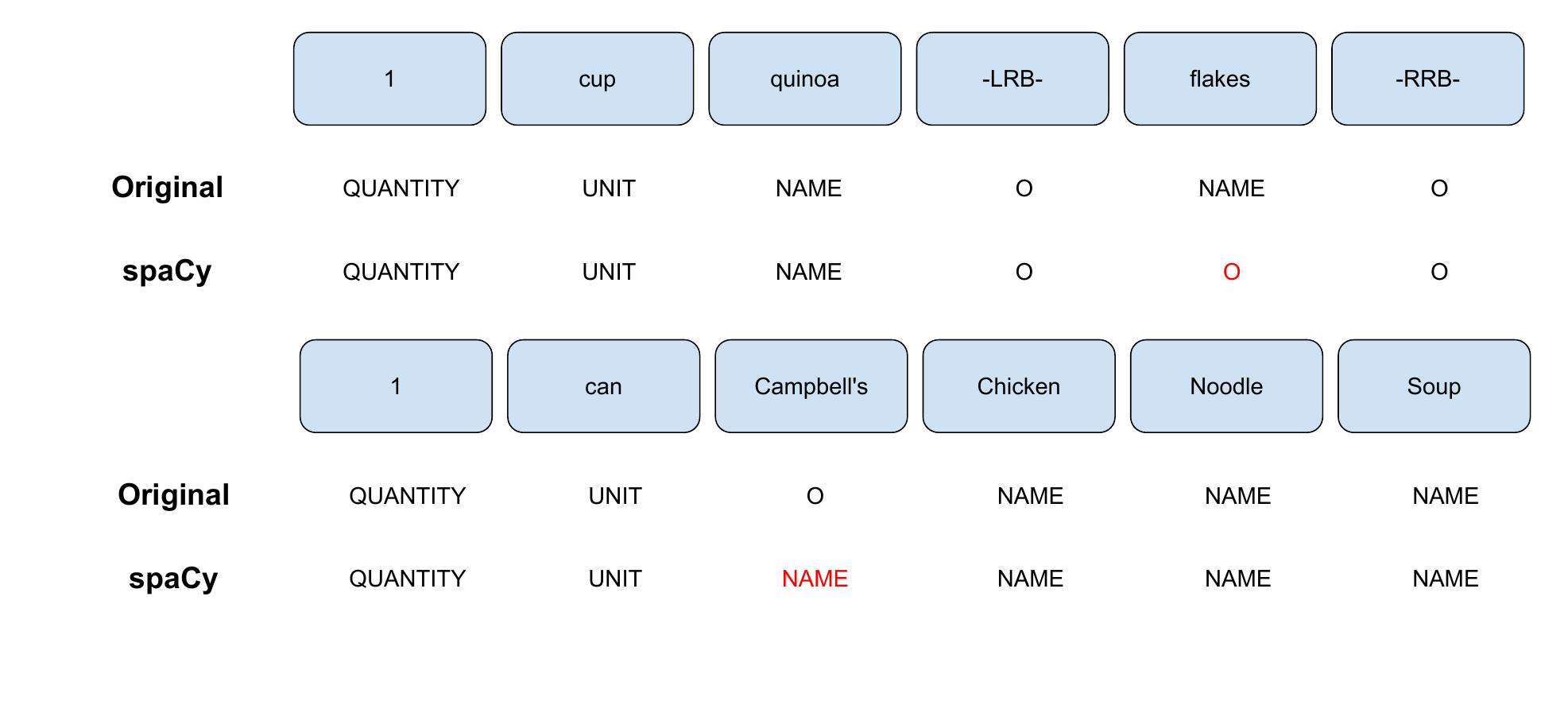} 
\caption{Error Analysis of spaCy-transformer. spaCy misclassifies ``flakes" as OTHER (O) instead of a specific form of quinoa (NAME). Similarly, ``Campbell's" is the chicken noodle soup brand name (O) instead of an ingredient name (NAME). -LRB- and -RRB- stand for left and right round brackets, respectively. }
\label{fig:spacy_error}
\end{center}
\end{figure*}

The spaCy-transformer model exhibits erroneous predictions, which include misclassification of ingredient names and brands (see Figure~\ref{fig:spacy_error}).

\section{Cleaning Protocols for Machine Annotated Dataset}
\label{sec:cleaning_appendix}

\begin{itemize}
    \item The model could not fully capture the unique culinary language dynamics different from our usual natural language. Color is an adjective, but it might be part of the ingredient. For example, ‘Yellow lentils’ where ‘yellow’ in natural language is a color and a usual natural language model would classify it as a ‘STATE’ of an ingredient. Other examples are red Romano pepper, red chilies, etc.

    \item Fixing the incorrect placement of named entities. These included the quantity incorrectly labeled as a unit or vice-versa and an ingredient incorrectly classified as a unit or vice-versa. A null value was used to indicate the absence of the unit in the unique list of training datasets. 

    \item Append the fraction and integer together in the quantity as a string to avoid misclassification.

    \item Removal of special characters. For example, in “+1 chikoo”, ‘+’ is dropped.
    
\end{itemize}

\end{document}